\pdfoutput=1

\documentclass[11pt]{article}

\usepackage{subfig}
\usepackage{amsmath}
\usepackage{booktabs}
\usepackage{multirow}
\usepackage{graphicx}
\usepackage{xcolor}
\usepackage{newfloat}
\usepackage{listings}
\usepackage{url}
\usepackage{bm}
\usepackage{colortbl}

\usepackage[final]{acl}

\usepackage{times}
\usepackage{latexsym}

\usepackage[T1]{fontenc}

\usepackage[utf8]{inputenc}

\usepackage{microtype}

\usepackage{inconsolata}

\usepackage{graphicx}

\title{Cross-Image Contrastive Decoding: Precise, Lossless Suppression of Language Priors in Large Vision-Language Models}

\author{
 \textbf{Jianfei Zhao\textsuperscript{1,2}},
 \textbf{Feng Zhang\textsuperscript{1}},
 \textbf{Xin Sun\textsuperscript{1}},
 \textbf{Lingxing Kong\textsuperscript{4}},
 \textbf{Zhixing Tan\textsuperscript{4}},
 \textbf{Chong Feng\textsuperscript{1,3}}
\\
 \textsuperscript{1}School of Computer Science and Technology, Beijing Institute of Technology\\
 \textsuperscript{2}Zhongguancun Academy\\
 \textsuperscript{3}Southeast Academy of Information Technology, Beijing Institute of Technology\\
\textsuperscript{4}Tsinghua University\\
 \small{
      \{zhqingan, bit\_zhangfeng, sunxin\}@bit.edu.cn, 
    klxstar@126.com,
    tzx.2019@tsinghua.org.cn,
    fengchong@bit.edu.cn
 }
}

\begin{document}

\maketitle
\begin{abstract}
Over-reliance on language priors is a major cause of hallucinations in Large Vision-Language Models (LVLMs), often leading to outputs that are linguistically plausible but visually inconsistent. Recent studies have explored contrastive decoding as a training-free solution. However, these methods typically construct contrastive visual inputs by perturbing the original image, resulting in distorted contrastive distributions, incomplete contrastive signals, and excessive suppression of language priors. Motivated by the observation that language priors tend to remain consistent across different images, we propose \textbf{C}ross-\textbf{I}mage \textbf{C}ontrastive \textbf{D}ecoding (CICD), a simple yet effective training-free method that uses unrelated images as contrastive visual inputs. To address the issue of over-suppressing language priors, which can negatively affect the quality of generated responses, we further introduce a dynamic selection mechanism based on the cross-image differences in model behavior. By selectively suppressing language priors, our method reduces hallucinations without compromising the model's performance. Extensive experiments across multiple benchmarks and LVLMs confirm the effectiveness and generalizability of CICD, particularly in image captioning, where language priors are especially dominant.
\footnote{Our code is available at \url{https://github.com/beta-nlp/CICD}.}
\end{abstract}

\section{Introduction}
\label{sec:intro}

Recent developments in Large Language Models (LLMs)~\cite{chiang2023vicuna,bai2023qwen} have opened a promising pathway toward Artificial General Intelligence (AGI). Building on these advances, many studies~\cite{liu2023visual,blip,qwen-vl} have focused on integrating visual inputs into LLMs, giving rise to Large Vision-Language Models (LVLMs) capable of understanding visual content. These models can handle a wide range of vision-language tasks, including image captioning~\cite{chair,DetailCap, chen2025perturbollava} and visual question answering (VQA)~\cite{POPE,lovenia2024negative,yu2024mm}.

\begin{figure}[tbp]
  \centering
  \includegraphics[width=\linewidth]{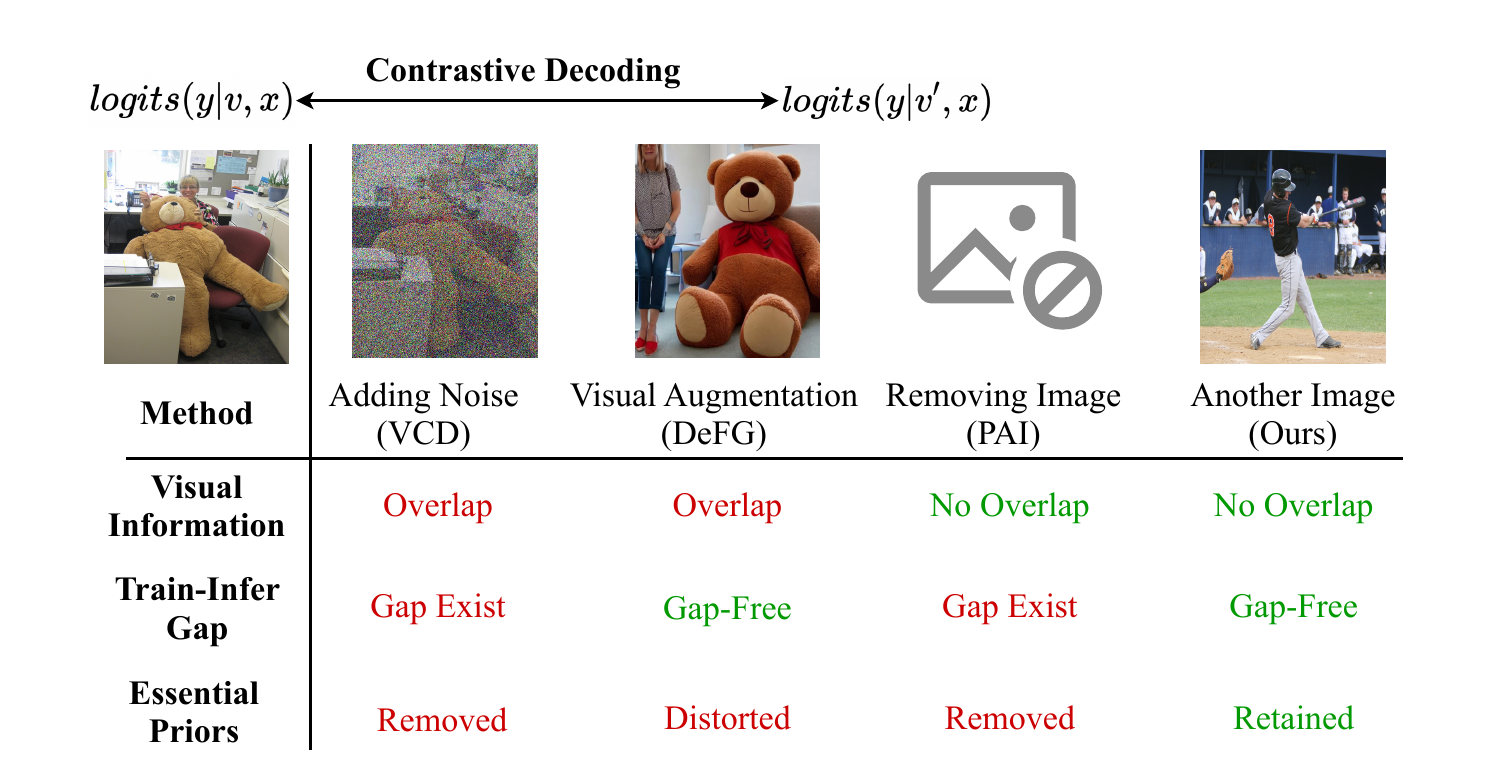}
  \caption{Comparison of contrastive decoding–based methods. Our approach avoids overlapping visual information, training-inference gap, and excessive suppression of language priors observed in existing methods.}
  \label{fig:overview}
\end{figure}

Despite their impressive capabilities, LVLMs are known to suffer from severe hallucinations~\cite{POPE,wang2023evaluation,VCD}. These models often generate responses based solely on memorized linguistic associations rather than actual visual input. For example, when describing an office scene, an LVLM may erroneously mention a “computer” even if no such object is present in the image. Therefore, improving the reliability and applicability of LVLMs across domains requires addressing this hallucination problem. While the causes of hallucination in LVLMs are complex, prior studies~\cite{VCD,ICD,IBD,chen2025perturbollava} have identified over-reliance on language priors as one of the primary contributing factors. Recent efforts to mitigate this issue have focused on plug-and-play, training-free methods~\cite{ICD,VCD,IBD,DeGF}, which avoid the substantial computational costs associated with training-based approaches~\cite{wu2022overcoming,ren2023overcoming,chen2025perturbollava}. These methods~\cite{VCD,ICD,IBD} commonly employ contrastive decoding~\cite{li2023contrastive}, which highlights differences in output distributions between the original and contrastive inputs.

Various strategies~\cite{VCD,ICD,PAI,favero2024multi,DeGF} have been proposed for constructing visual inputs in contrastive decoding. While previous studies have demonstrated their effectiveness, we identify three major challenges in contrastive decoding-based methods, as illustrated in Figure~\ref{fig:overview}. First, \emph{distorted distributions from contrastive visual inputs}. \citet{VCD,ICD,PAI,favero2024multi} propose using noised images or even removing the image entirely when computing the contrastive distribution. However, these approaches introduce a training-inference gap, as such contrastive inputs are not seen during training. This may result in distorted contrastive distributions. Second, \emph{overlapping visual information between original and contrastive inputs}. \citet{VCD,ICD,DeGF} propose using noisy or generated images that share visual similarities with the original inputs. However, such strategies may lead to incomplete contrastive signals. Third, \emph{trade-offs in suppressing language priors}. Although over-reliance on language priors can cause hallucinations, excessively suppressing them may degrade the performance of the model. Although \citet{DeGF} selectively remove language priors, they augment the retained language priors, which introduces bias. As a result, mitigating language priors through contrastive decoding remains an open challenge that needs further investigation.

\begin{figure}[tbp]
  \centering
  \includegraphics[width=\linewidth]{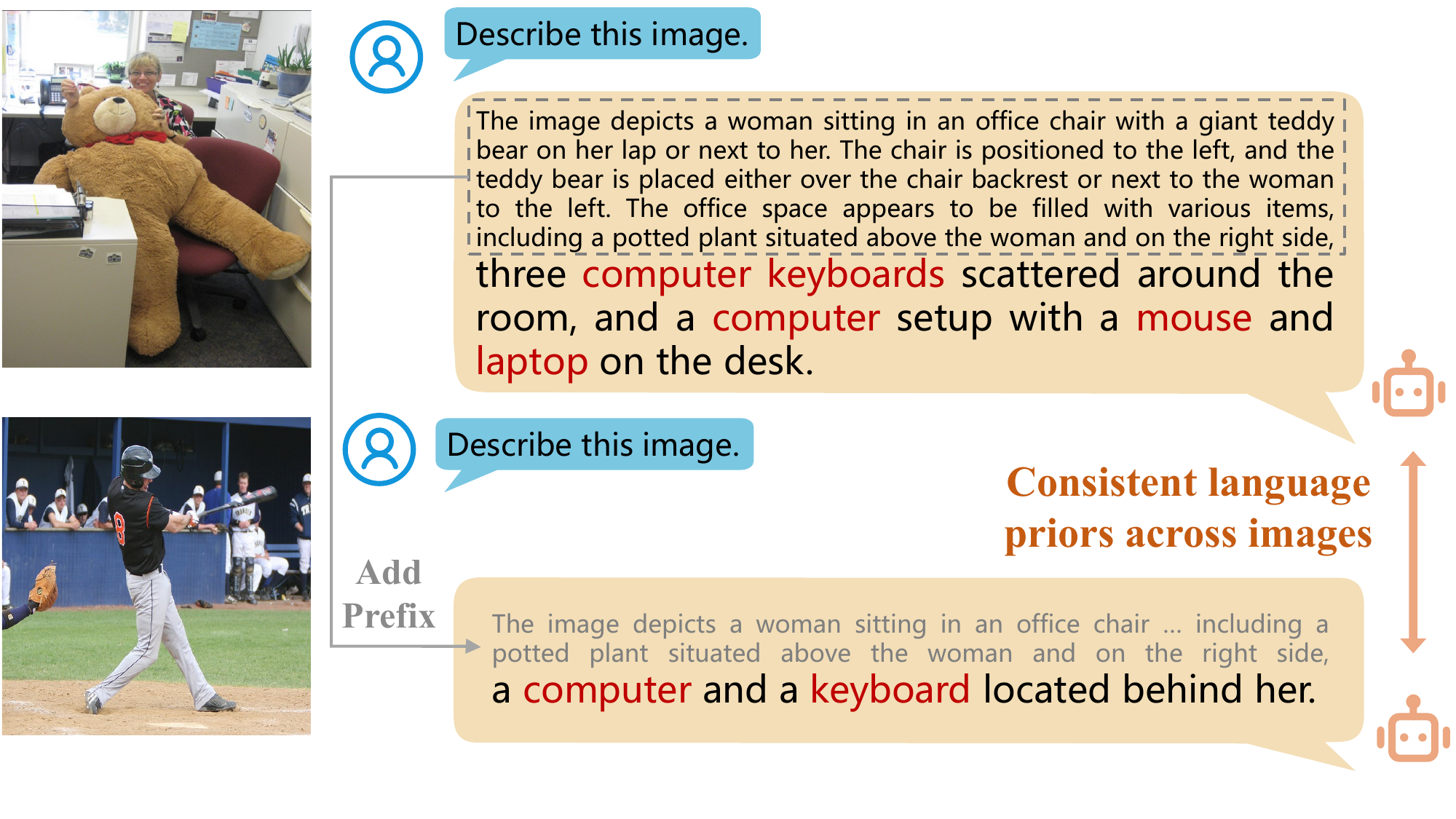}
  \caption{
  Adding the same prefix when captioning a different image leads to the same hallucination, indicating that language priors are consistent across images.
  }
  \label{fig:case}
\end{figure}

In this paper, we propose Cross-Image Contrastive Decoding (CICD), a simple yet effective method to address the aforementioned challenges in contrastive decoding. We begin by observing that LVLMs often produce highly similar outputs when given the same textual prefix but entirely different images, suggesting that language priors tend to dominate regardless of visual input (see Figure~\ref{fig:case} for an example). Based on this observation, we construct contrastive visual inputs using different images. This approach avoids the training-inference gap and minimizes overlap between the original and contrastive inputs, thereby emphasizing the visual information in the original input and alleviating over-reliance on language priors. To prevent excessive suppression of language priors during contrastive decoding,
we leverage the consistency of model behavior across images to dynamically determine, at each decoding step, whether language priors should be suppressed and to what extent.
When the divergence falls below a predefined threshold, our method backs off to regular decoding. This adaptive mechanism allows CICD to effectively reduce over-reliance on language priors while preserving the model's performance.

We conduct extensive experiments across multiple benchmarks and LVLMs. The results demonstrate that our method effectively alleviates language priors, particularly in the image captioning task.

\section{Related Work} \label{sec:related work}

\subsection{Hallucination in LVLMs}
In the multimodal domain, the study of hallucinations primarily focuses on cross-modal consistency~\cite{POPE,wang2023evaluation}, which requires that content generated by LVLMs remains faithful to the visual input. Multiple factors contribute to hallucinations in LVLMs, including biased visual attention~\cite{DAC,IMCCD}, attention misallocation~\cite{opera,VAF,VAR}, and over-reliance on language priors~\cite{VCD, IBD, ICD}. Our work primarily focuses on mitigating the over-reliance on language priors in LVLMs.

\subsection{Language Priors in LVLMs}
As LVLMs are built upon LLMs, they inherently inherit the linguistic knowledge embedded in LLMs—commonly referred to as language priors~\cite{wu2022overcoming,ren2023overcoming}. Over-reliance on these priors can cause LVLMs to generate content that is linguistically plausible but inconsistent with visual evidence. To mitigate this issue, \citet{liu2024mitigating,gunjal2024detecting,chen2025perturbollava} design specialized training datasets. However, these training-based approaches are often resource-intensive and not easily transferable to broader application scenarios. Recent studies have explored contrastive decoding~\cite{li2023contrastive,chuang2023dola} as a training-free alternative to reduce dependence on language priors. These methods construct an alternative logit distribution alongside the original one, and then contrast the two distributions to suppress language priors during decoding. The contrastive distribution is typically generated using techniques such as masking the image~\cite{VCD}, perturbing the instruction~\cite{ICD}, augmenting the visual input~\cite{IBD}, or performing cross-modal conversion~\cite{DeGF}. Our approach also builds on the contrastive decoding framework. However, unlike the aforementioned methods, we propose constructing contrastive inputs using a different image and introducing a dynamic mechanism to prevent the over-suppression of language priors. This design avoids both the introduction of overlapping visual information and the training-inference discrepancies in the contrastive input, making it more effective at addressing hallucinations in LVLMs.

\section{Cross-Image Contrastive Decoding}
\begin{figure*}[tbp]
  \centering 
  \includegraphics[width=1\linewidth]{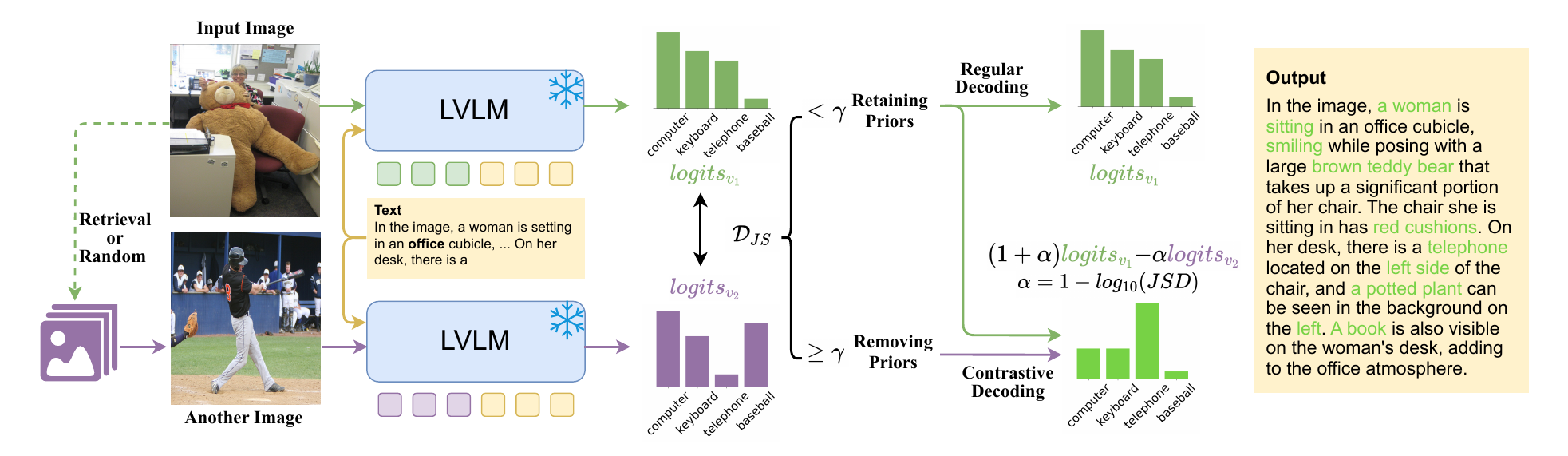}
  \caption{\label{Fg_Diagram}
  Overview of our method. We input the target image and a distinct image into the LVLM separately. During step-by-step response generation, we compute the Jensen–Shannon (JS) divergence between the two logit distributions at each step to decide whether to suppress language priors. Contrastive decoding is applied to suppress language priors, while regular decoding is used to avoid excessive suppression.
  }
\end{figure*}

We propose Cross-Image Contrastive Decoding (CICD) for effective mitigation of language priors. CICD first retrieves a distinct image as the contrastive visual input. Then, at each decoding step, it dynamically chooses between regular decoding and contrastive decoding based on the divergence between the original and contrastive distributions. An overview of CICD is presented in Figure~\ref{Fg_Diagram}. We begin with an analysis of language priors in Section~\ref{sec:lp_analysis}. Based on this analysis, we present methods for constructing contrastive visual inputs in Section~\ref{sec:img_retrieval}. Finally, we discuss the importance of balanced language prior suppression in Section~\ref{sec:dyn_supp}.

\subsection{An Analysis of Language Priors} \label{sec:lp_analysis}

\begin{figure}[tbp]
  \centering
  \includegraphics[width=\linewidth]{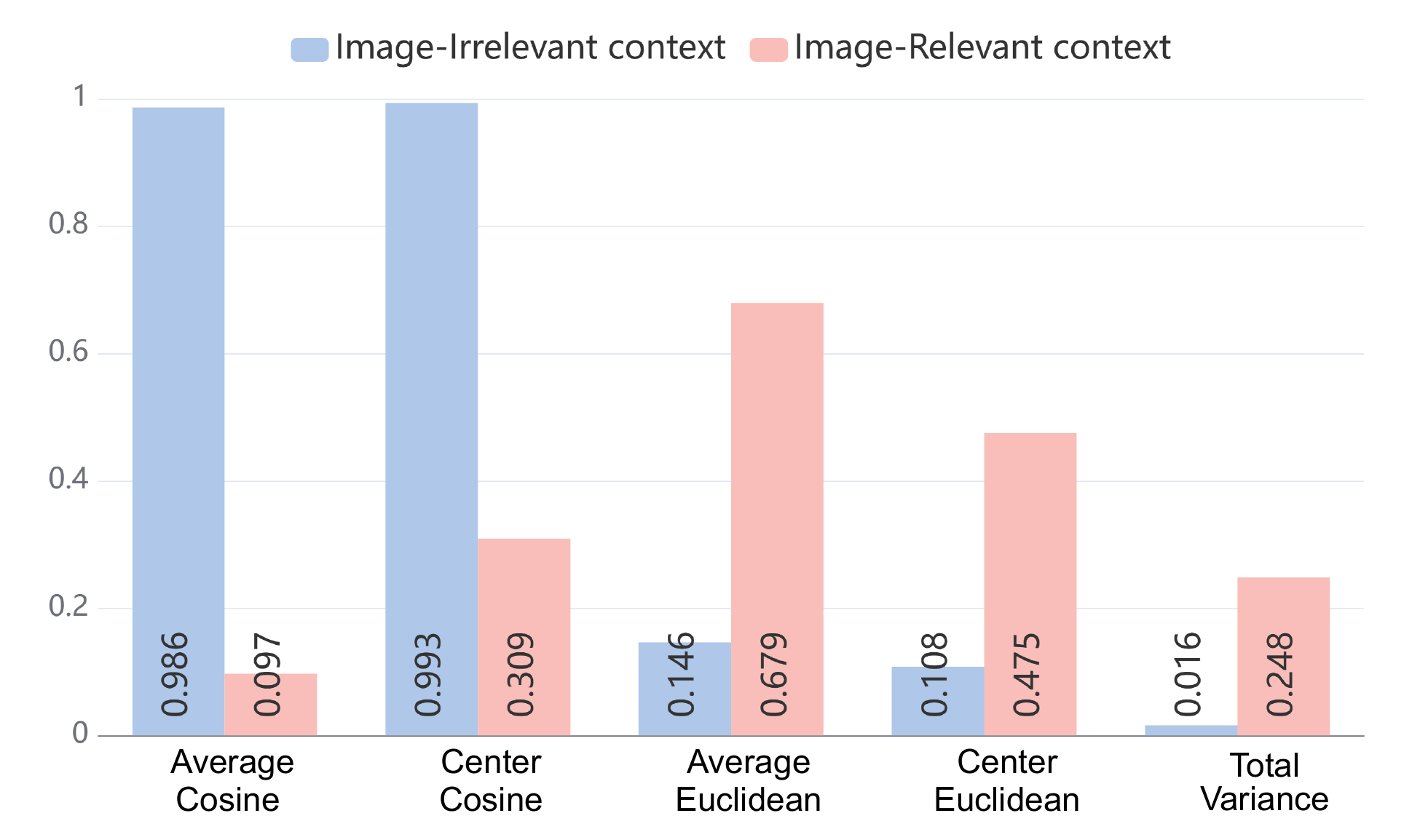}
  \caption{\label{fig:consistency}
    Similarity of the output logit distributions of LLaVA-1.5-7B in image-relevant and image-irrelevant contexts across 500 images.
  }
\end{figure}

While over-reliance on language priors may lead to hallucinations in LVLMs, excessive suppression of these priors can negatively affect the quality of generated texts. To investigate the effect of language priors on the generation of image-relevant and image-irrelevant tokens, we examine the logit distributions of LVLMs in both image-relevant and image-irrelevant contexts.

To this end, we randomly selected 500 images from the MSCOCO dataset and compared the similarity of the output logit distributions produced by LLaVA-1.5-7B in image-relevant and image-irrelevant contexts.
We define the generation for the first token as the image-irrelevant context, and use the prefix ``\emph{The picture depicts a}'' as the image-relevant context.
We use cosine distance, Euclidean distance, and variation to study the similarity between logit distributions in the image-irrelevant and image-relevant contexts, respectively.
It is evident that the logit distributions in image-irrelevant contexts exhibit high similarity across different images, regardless of the similarity metric used. In contrast, the logit distributions in image-relevant contexts do not display such consistency. We further validate this observation using additional context prompts, and the results remain consistent,
suggesting that cross-image similarity can be leveraged to discriminate between image-irrelevant and image-relevant contexts.
Only language priors in image-relevant contexts need to be removed.

\subsection{Construction of Contrastive Visual Inputs} \label{sec:img_retrieval}

Based on the above analysis, we propose constructing contrastive visual inputs using an image that is distinct from the original input image. By choosing an image from a given dataset instead of synthesizing one, we avoid introducing a potential training–inference gap while maximizing visual contrast during contrastive decoding.

Intuitively, the image with the lowest similarity to the input image is preferred as the contrastive visual input for contrastive decoding. We propose two different methods for selecting a distinct image from a given dataset $\mathcal{D}_{\textrm{img}}$:
\paragraph{Retrieval.} Since visual features are extracted by the vision encoder, a natural strategy is to measure the similarity between two images using the encoder itself to estimate their overlap. To ensure better generalizability, we adopt the pretrained CLIP model~\cite{CLIP} as the retriever. As all images in the dataset $\mathcal{D}_{\mathrm{img}}$ can be vectorized in advance before the inference stage, the retrieval process incurs negligible computational cost. We simply retrieve the image with the lowest cosine similarity to the input, which is then used as the contrastive image during decoding.
\paragraph{Random.} Although vector retrieval is relatively efficient compared to the computational cost of LVLMs, practical deployment still requires loading an additional retriever and maintaining a vector database. Previous works~\cite{POPE,AMBER,DetailCap} reveal that language priors primarily exist in content words such as objects, attributes, and relationships. These visual elements tend to exhibit minimal overlap across natural images. Therefore, randomly selecting an image can serve as an effective and lightweight alternative for practical applications.

\subsection{Dynamic Language Prior Suppression} \label{sec:dyn_supp}
Although language priors may cause LVLMs to generate hallucinated content, they are essential for the model's generation capabilities. As previously analyzed, language priors exhibit high consistency across different images in image-irrelevant contexts. Therefore, we employ the Jensen–Shannon (JS) divergence to measure the consistency between two logit distributions produced from different images $\bm{v}$ and $\bm{v}'$, which is formally defined as
\begin{equation}
d(\bm{v},\bm{v}') = \mathrm{JSD}(p_{\bm{\theta}}(y_t | \bm{v}, \mathbf{x}, \mathbf{y}_{<t})\ \|\  p_{\bm{\theta}}(y_t | \bm{v}', \mathbf{x}, \mathbf{y}_{<t}),
\end{equation}
where $\mathrm{JSD}(P\|Q) = \frac{1}{2}\mathrm{KL}(P\|M) + \frac{1}{2}\mathrm{KL}(Q\|M)$, $M = \frac{1}{2}(P + Q)$, $\mathrm{KL}(P\|Q)$ denotes the KL divergence between two probability distributions $P$ and $Q$, and $p_{\bm{\theta}}(y_t \mid \bm{v}, \mathbf{x}, \mathbf{y}_{<t})$ is the probability distribution of the next token $y_t$ given visual input $\bm{v}$, textual input $\mathbf{x}$, and the previously generated token sequence $\mathbf{y}_{<t}$. A higher value of $d(\bm{v}, \bm{v}')$ indicates that the current context is more image-relevant, while a lower value may suggest an image-irrelevant generation.

\begin{figure}[tbp]
  \centering
  \includegraphics[width=\linewidth]{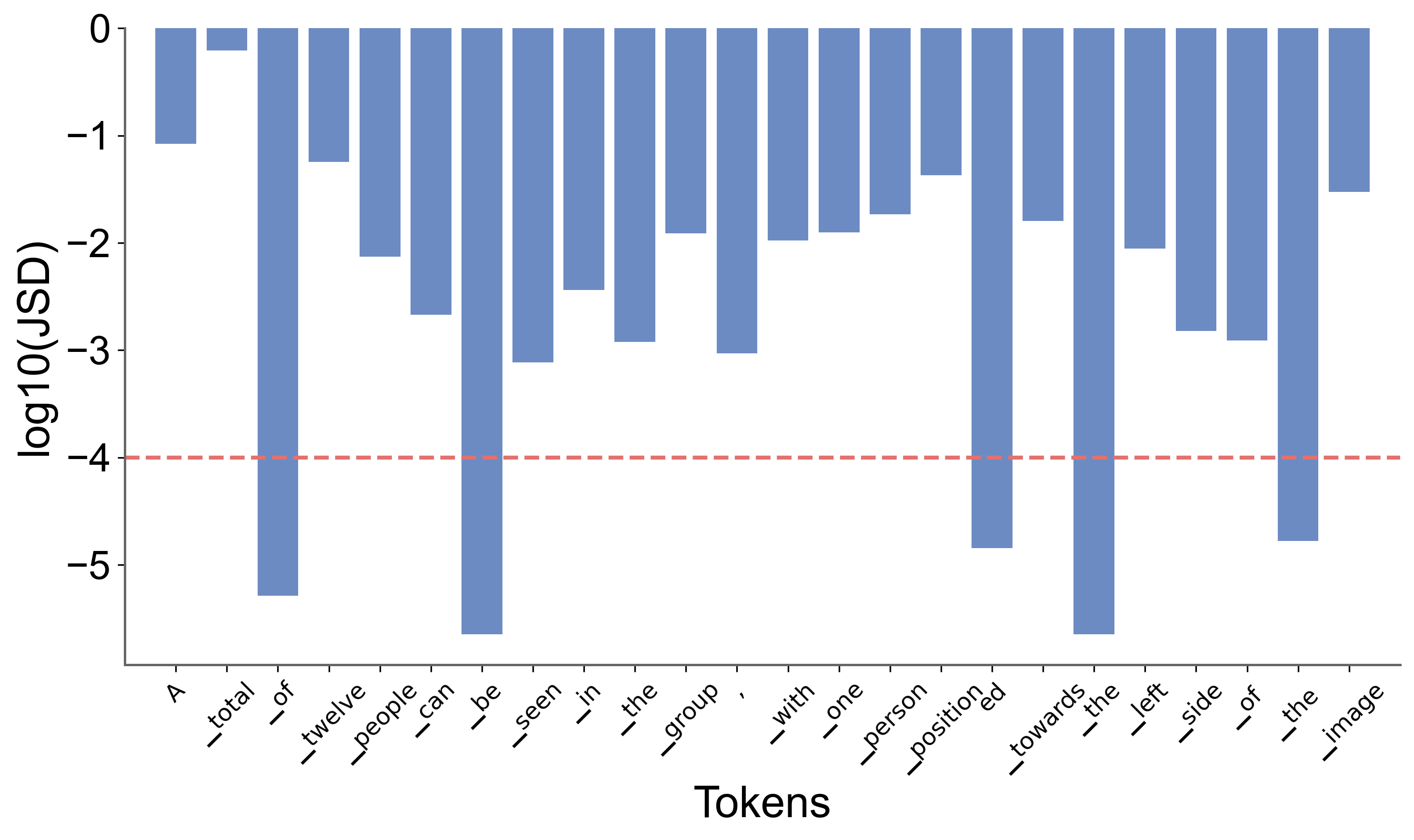}
  \caption{
    JS divergence between the logit distributions of two distinct images.
  }
  \label{fig:consisitency}
\end{figure}

\begin{figure}
  \centering
  \includegraphics[width=0.8\linewidth]{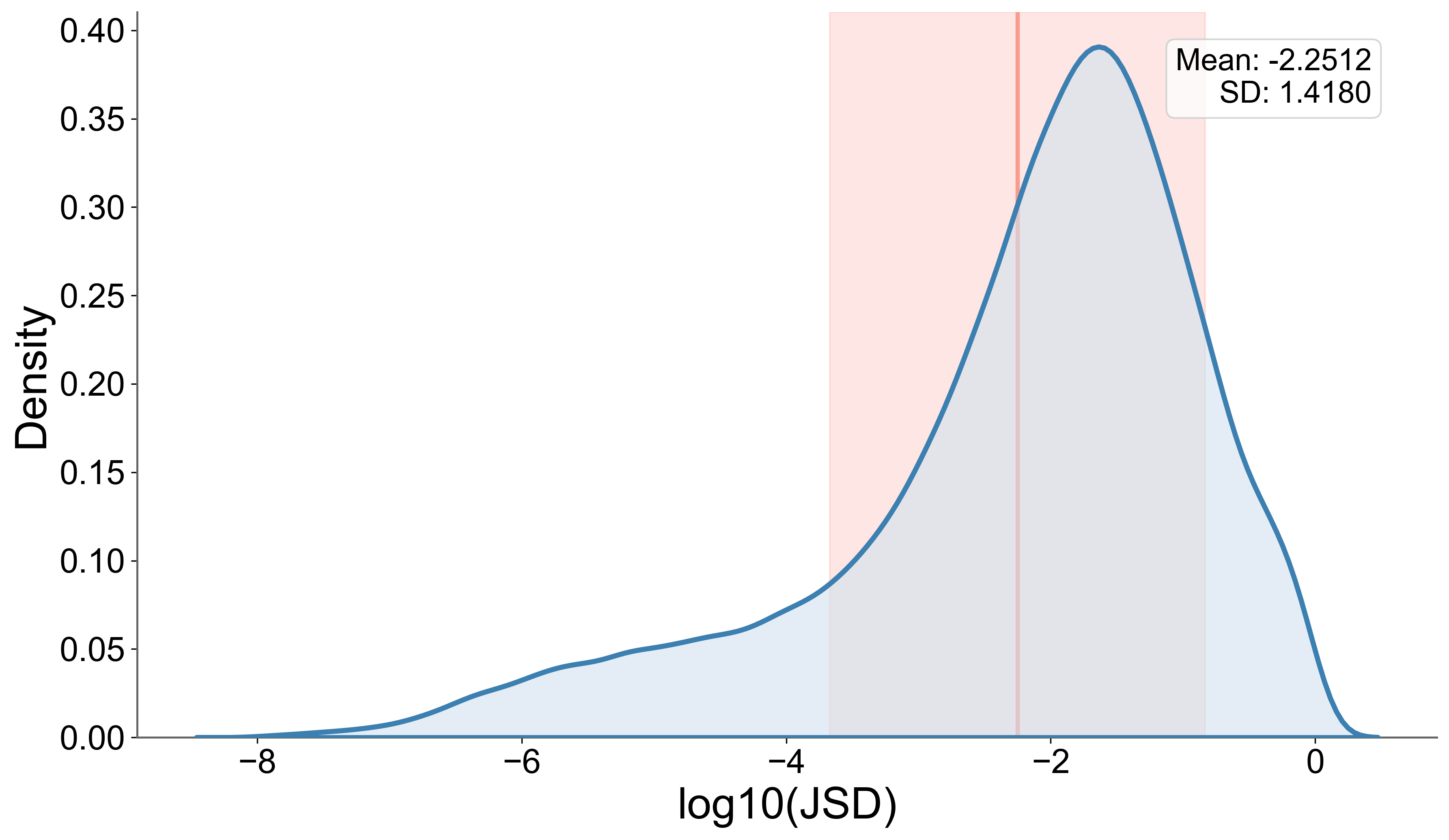}
  \caption{
    500 random images from the MSCOCO dataset captioned by LLaVA-1.5.
    For each next-token prediction, we statistic the JS divergence (JSD) between two logit distributions produced by different images.
  }
  \label{fig:jsd_kde}
\end{figure}

To demonstrate the effectiveness of JS divergence in determining context relevance, we plot the logarithm of the JS divergence between distributions generated with different images at each decoding step. Figure~\ref{fig:consisitency} shows the results. Clearly, image-relevant tokens such as ``\emph{\_people}'', ``\emph{\_white}'', and ``\emph{\_red}'' exhibit higher divergence values (often above $-4$ when taking $\log_{10}(d(\bm{v}, \bm{v}'))$), while function words such as ``\emph{\_of}'', ``\emph{\_be}'' and ``\emph{\_the}'' have lower JS divergence values (often below $-4$ in $\log_{10}(d(\bm{v}, \bm{v}'))$). We further randomly select 500 images from the MSCOCO dataset, captioned by LLaVA-1.5, and compute the JS divergence between logit distributions produced by different images for each next-token prediction. The statistical results are visualized in Figure~\ref{fig:jsd_kde}. We observe that the JS divergence values also fall into two distinct ranges: $\log_{10}(d(\bm{v}, \bm{v}')) > -4$ and $\log_{10}(d(\bm{v}, \bm{v}')) \leq -4$. These results further confirm that JS divergence can effectively distinguish between image-relevant and image-irrelevant contexts.

Based on the above analysis, CICD enables dynamic suppression of language priors. Specifically, CICD performs regular decoding at steps where image-irrelevant contexts are detected, while applying contrastive decoding with different images to suppress language priors in image-relevant contexts. A hyperparameter $\gamma$ is used to determine whether the current decoding step is image-irrelevant. Formally, the final logit in CICD is calculated as
\begin{align}\label{Eq_6}
\mathrm{logit}(y_t) &= \left\{
\arraycolsep=1.0pt
    \begin{array}{ll}
      \mathrm{logit}_{\theta}(y_t\ |\ \bm{v}, \mathbf{x}, \mathbf{y}_{<t}), & \textrm{if } d(\bm{v}, \bm{v}') \leq \gamma \\
      \mathrm{logit}_{\textrm{CD}}(y_t \ |\ \bm{v}, \mathbf{x}, \mathbf{y}_{<t}), & \textrm{if } d(\bm{v}, \bm{v}') > \gamma
    \end{array}\right.
\end{align}
where $\mathrm{logit}_{\textrm{CD}}(y_t | \bm{v},\mathbf{x}, \mathbf{y}_{<t})$ is defined as
\begin{equation}
\begin{aligned}\label{Eq_2}
\mathrm{logit}_{\textrm{CD}}(y_t | \bm{v}, \mathbf{x}, \mathbf{y}_{<t}) 
= & (1 + \alpha)\textrm{logit}_{\bm{\theta}}(y_t | \bm{v}, \mathbf{x}, \mathbf{y}_{<t}) \\ &-\alpha\textrm{logit}_{\bm{\theta}}(y_t | \bm{v}', \mathbf{x}, \mathbf{y}_{<t}), 
\end{aligned}
\end{equation}
where $\alpha$ is the contrastive coefficient. Unlike previous approaches such as \citet{VCD}, $\alpha$ is not treated as a hyperparameter in our method. Instead, it is dynamically set to $1 - \log_{10}(d(\bm{v}, \bm{v}'))$.
This is based on the intuition that detrimental priors with higher similarity are more difficult to remove and therefore require stronger suppression. To ensure numerical stability, we clip $\alpha$ to the interval $[1, 3]$.




\begin{table*}[t]
  \centering
  \begin{tabular}{llcccccc}
    \toprule
    \multirow{2}{*}{\textbf{Setting}} & \multirow{2}{*}{\textbf{Method}} & \multicolumn{2}{c}{LLaVA-1.5} & \multicolumn{2}{c}{InstructBLIP} & \multicolumn{2}{c}{Qwen-VL-Chat}  \\
    \cmidrule(rl){3-4} \cmidrule(rl){5-6} \cmidrule(rl){7-8}
    &  & \textbf{Acc.}$\uparrow$ & \textbf{F1}$\uparrow$ & \textbf{Acc.}$\uparrow$ & \textbf{F1}$\uparrow$ & \textbf{Acc.}$\uparrow$ & \textbf{F1}$\uparrow$\\
     \midrule
     \multirow{7}{*}{Random} & Regular & 83.47 & 82.25 & 80.49 & 81.10 & 82.32 & 79.35  \\
                      & VCD~\cite{VCD} & 85.30 & 84.59 & 81.75 & 82.06 & \underline{84.74} & \underline{82.80} \\
                     & ICD~\cite{ICD} &  84.81 & 83.44 & 80.95 &81.59  & 83.10 & 80.52  \\
                     & PAI~\cite{PAI} &   84.56 & 83.33 & 79.54 & 80.27 &  84.14 & 81.77  \\
                     & IBD~\cite{IBD} &   85.36 & 84.28 &74.82 & 77.87 & 83.07 & 80.23    \\
                      & DeGF~\cite{DeGF} & \underline{86.68} &\underline{85.71} &  \textbf{84.35} &\underline{84.50} &  84.04 &81.49    \\
                      \rowcolor{gray!20}
                      & CICD (\emph{ours}) &  \textbf{87.85} &\textbf{87.77} &\underline{84.00} &\textbf{84.89}& \textbf{87.54} &\textbf{86.37}  \\
    \midrule
    \multirow{7}{*}{Popular} & Regular & 79.56 &78.99 &76.59 &78.16 &81.11 &78.34     \\
                            & VCD~\cite{VCD}  &  80.81 & 80.84 & 77.22 & 78.60 & 82.95 & \underline{81.17}                             \\
                            & ICD~\cite{ICD} &  81.44 & 80.53 & 76.82 & 78.48 & 81.26 &78.87                             \\
                            & PAI~\cite{PAI} &  80.27 & 79.72 & 75.04 & 77.05 & 83.06 & 80.89                          \\
                            & IBD~\cite{IBD} &  81.68 & 81.13 & 69.35 & 74.31 & 82.20 &79.59                             \\
                            & DeGF~\cite{DeGF} &  \underline{82.96} &\underline{82.46} &\underline{79.25} &\underline{80.46} &\underline{83.02} &80.57                             \\
                      \rowcolor{gray!20}
                      & CICD (\emph{ours}) & \textbf{84.83} &\textbf{85.18}&\textbf{81.01} &\textbf{82.53} &\textbf{85.94} &\textbf{84.90}                             \\
    \midrule
    \multirow{7}{*}{Adversarial} & Regular & 76.95 &77.13 & 72.89 &75.58 &78.23 &75.72     \\
                            & VCD~\cite{VCD}  & 77.27 & 78.19 & 74.03 & 76.39 & 80.37 &\underline{78.96}                             \\
                            & ICD~\cite{ICD} &   78.07 &77.98& 73.19 &75.91& 79.03 &77.05                             \\
                            & PAI~\cite{PAI} &    77.02 & 77.30 & 72.14 & 75.15 &\underline{80.62} & 78.76                         \\
                            & IBD~\cite{IBD} &   78.26 &78.52& 66.60 &72.63 &79.68 &77.36                             \\
                            & DeGF~\cite{DeGF} &  \textbf{79.40} &\underline{79.69} &\underline{75.19} &\underline{77.52} &80.42 &78.24                             \\
                      \rowcolor{gray!20}
                      & CICD (\emph{ours}) & \underline{78.60} &\textbf{80.30} &\textbf{75.55} &\textbf{78.63} &\textbf{81.94} & \textbf{81.37}                             \\
    \bottomrule
  \end{tabular}
  \caption{
      Results on POPE.
      \textbf{Acc.} and \textbf{F1} represent the average accuracy and F1-score across three datasets (MSCOCO, A-OKVQA, and GQA), respectively.
      The \textbf{bolded} results denote the best in each setting, and the \underline{underlined} results denote the second-best.
    }\label{tab:pope}
\end{table*}



\section{Experiment}

\subsection{Setup}
\paragraph{Benchmarks.}
We evaluate our CICD on 4 widely adopted multimodal hallucination benchmarks: 
\textbf{DetailCaps}~\cite{DetailCap}: A fine-grained image captioning benchmark. We evaluate on a 700-image MSCOCO subset.  
\textbf{POPE}~\cite{POPE}: A hallucination evaluation benchmark with 9,000 questions and 1,500 images from \emph{MSCOCO}~\cite{coco}, \emph{A-OKVQA}~\cite{aokvqa}, and \emph{GQA}~\cite{gqa}, covering three negative sampling settings (\emph{Random}, \emph{Popular}, and \emph{Adversarial}).
We also evaluate our method on the \textbf{CHAIR}~\cite{chair} and \textbf{AMBER}~\cite{AMBER} benchmarks. Due to space constraints, we leave the results in Appendix~\ref{sec:Detailed Experiment}.

\paragraph{LVLMs.} We use 6 representative models to study the effectiveness of CICD, which are \textbf{InstructBLIP}~\cite{blip}, \textbf{LLaVA-1.5}~\cite{llava15}, \textbf{LLaVA-Next}~\cite{llava-next}, \textbf{Qwen-VL-Chat}~\cite{qwen-vl}, \textbf{Qwen2-VL-Instruct}~\cite{qwen2-vl}, and \textbf{Qwen2.5-VL-Instruct}~\cite{qwen25-vl} respectively. The parameter counts of the above models range from 7B to 8B unless otherwise stated.

\paragraph{Baselines.}
We compare CICD with the following representative training-free methods: 
\textbf{VCD}~\cite{VCD}: a method that introduces noise into the input image during contrastive decoding to suppress language priors.
\textbf{ICD}~\cite{ICD}: a method that also uses contrastive decoding but adopts adversarial instruction to construct negative logits.
\textbf{IBD}~\cite{IBD}: a method that increases the influence of image information during decoding through strengthening the model's attention to visual features.
\textbf{PAI}~\cite{PAI}: a method that removes the visual input from the original input and retains the pure textual context as the negative context, while simultaneously enhancing the model's attention to visual tokens in the original context.
\textbf{DeGF}~\cite{DeGF}: a method that converts the generated caption into an image and re-generates the caption from that image, leveraging the preserved visual content to reduce hallucinations.
The implementation details of baselines are provided in the Appendix~\ref{sec:Implementation Details for Baselines}.

\paragraph{Implementation Details.}
We use sampling decoding for next-token prediction with default settings.
Following \citet{VCD}, we incorporate adaptive plausibility constraints into the decoding process.
For image retrieval, we use the MSCOCO 2014 validation set as $\mathcal{D}_{\textrm{img}}$, which contains 40,504 images.
All experiments are conducted on a single NVIDIA RTX 3090 GPU. Unless otherwise noted, we set the threshold $\gamma=-4$ in all our experiments.

\subsection{Main Results}
\begin{table*}[t]
  \centering
  \begin{tabular}{lccccccccc}
    \toprule
     \multirow{2}{*}{\textbf{Method}} & \multicolumn{3}{c}{\textbf{LLaVA-1.5}} & \multicolumn{3}{c}{\textbf{InstructBLIP}} & \multicolumn{3}{c}{\textbf{Qwen-VL-Chat}}  \\
    \cmidrule(rl){2-4} \cmidrule(rl){5-7}  \cmidrule(rl){8-10} 
     & \textbf{Cs} $\downarrow$ & \textbf{Ci}$\downarrow$ & \textbf{CAP}$\uparrow$ & \textbf{Cs}$\downarrow$ & \textbf{Ci}$\downarrow$ & \textbf{CAP}$\uparrow$ & \textbf{Cs}$\downarrow$& \textbf{Ci}$\downarrow$ & \textbf{CAP}$\uparrow$ \\
     \midrule
      Regular & 55.7&17.4 & 52.60 & 58.6&18.0 &  52.99  & 9.9&7.3 &30.73       \\
      VCD~\cite{VCD}  &  55.7&16.8 &\underline{52.91}  & 59.6&18.9 & 53.20  & 7.3&5.9 &\underline{31.28}       \\
      ICD~\cite{ICD} &  53.9&16.6 &52.82 & \underline{55.7}&16.7  & 53.24   & 5.6&7.0 &29.70       \\
      PAI~\cite{PAI} &  \textbf{39.4} & \textbf{11.8} & 53.49  & 62.6 & 18.3 & 53.27  &  \textbf{4.6} & \underline{3.3} & 29.54       \\
      IBD~\cite{IBD} &  54.1&15.8 & 52.48 & 56.4&\underline{15.5} & \underline{54.14} & \textbf{4.6}&\textbf{2.9} & 29.31       \\
      DeGF~\cite{DeGF} & 55.7&16.6 & 52.72 & 59.1&17.4 & 53.06 & 5.6&3.9 & 30.21       \\
      \rowcolor{gray!20}
      CICD (\emph{ours}) & \underline{45.6}&\underline{13.1} &\textbf{55.80} & \textbf{45.7}&\textbf{13.2} & \textbf{54.20} & 5.9&3.5 & \textbf{31.34}       \\
    \bottomrule
  \end{tabular}
  \caption{
      Results on the COCO subset of DetailCaps.
      \emph{Cs}, \emph{Ci}, and \emph{CAP} denote
      \emph{CHAIRs}, \emph{CHAIRi}, and 
\emph{CAPTURE} respectively.
      }\label{tab:dc}
\end{table*}

\paragraph{Results on POPE.}
Table~\ref{tab:pope} presents the results of all baseline methods alongside our proposed method. Our approach achieves the highest F1 scores across all sampling settings and backbone LVLMs. These results suggest that CICD effectively mitigates hallucinations and promotes the use of faithful visual information in LVLMs. Notably, under the \emph{Popular} setting, which uses frequent objects as negative samples and thus increases the likelihood of over-reliance on language priors. Nevertheless, our CICD method still achieves significant improvements. This highlights its effectiveness in reducing dependency on language priors.

\paragraph{Results on DetailCaps.}
Table~\ref{tab:dc} presents the results on the MSCOCO subset of DetailCaps. We primarily use \emph{CAPTURE} as the main evaluation metric, which assesses caption quality through both lexical matching and semantic alignment by measuring the consistency of key content (entities, attributes, and relations) between generated captions and multiple ground-truth references. Our CICD method achieves the best performance across all three LVLMs, demonstrating strong image captioning capabilities and confirming its generalizability across different datasets. These results suggest that CICD is effective in mitigating the over-reliance on language priors.

\subsection{Comparison of Retrieval Methods}\label{sec:Similarity Between Images}
\begin{table}[tbp]
  \centering
  \setlength{\tabcolsep}{1mm}
  \small{
    \begin{tabular}{lcccccc}
    \toprule
    \multirow{2}{*}{\textbf{Method}} & \multicolumn{3}{c}{\textbf{CHAIR}} & \multicolumn{3}{c}{\textbf{POPE} (F1$\uparrow$)}   \\
    \cmidrule(rl){2-4}  \cmidrule(rl){5-7}
    & \textbf{Cs}$\downarrow$ & \textbf{Ci}$\downarrow$ & \textbf{R} & \textbf{Ran}& \textbf{Pop} & \textbf{Adv}   \\
    \midrule
    Regular & 54.6&16.4&72.6 & 82.25 & 78.99 & 77.13 \\
    CICD$_{\textrm{Random}}$ & 46.6 & 13.2 & \textbf{76.7} & \textbf{87.86} & 85.13 & 80.16  \\
    CICD$_{\textrm{Retrieved}}$ & \textbf{43.8} & \textbf{11.7} & 75.0 & 87.77 & \textbf{85.18} & \textbf{80.30} \\
    \bottomrule
  \end{tabular}
}
  \caption{\label{tab:img}
  Comparison of CICD with different contrastive input construction methods on the CHAIR and POPE benchmarks. \emph{Ran}, \emph{Pop}, and \emph{Adv} denote the Random, Popular, and Adversarial subsets, respectively.}
\end{table}

We study the performance of CICD using different contrastive input construction methods. To this end, we conduct experiments on the CHAIR and POPE benchmarks using LLaVA-1.5 as the backbone LVLM. We denote CICD with random sampling as CICD$_{\textrm{Random}}$ and CICD with image retrieval as CICD$_{\textrm{Retrieved}}$. The results are presented in Table~\ref{tab:img}. It is evident that both CICD$_{\textrm{Random}}$ and CICD$_{\textrm{Retrieved}}$ significantly outperform regular decoding on both benchmarks.

We further analyze the similarity between the original and contrastive visual inputs by computing the overlap ratio of image-related words in their descriptions within the COCO dataset.
We find that the overlap ratios for random sampling and image retrieval are only 0.24\% and 0.11\%, respectively, relative to the original inputs, which are significantly lower than those of other methods, as shown in Figure~\ref{fig:vis_sim}.
These findings suggest that using a visually distinct image is effective for contrastive decoding. Moreover, random sampling can achieve performance comparable to image retrieval, provided that the dataset $\mathcal{D}_{\textrm{img}}$ contains sufficiently diverse images.

\subsection{Analysis}

\begin{table}[t]
  \centering
    \begin{tabular}{lcccc}
    \toprule
    \multirow{2}{*}{\textbf{Method}} & \multicolumn{2}{c}{\textbf{CHAIR$_{64}$}} & \multicolumn{2}{c}{\textbf{CHAIR$_{512}$}}   \\
    \cmidrule(rl){2-3}  \cmidrule(rl){4-5}
    & \textbf{Cs}$\downarrow$ & \textbf{Ci}$\downarrow$ & \textbf{Cs}$\downarrow$ & \textbf{Ci}$\downarrow$  \\
    \midrule

    PAI  & 20.0 & 6.2 & 41.2 & \textbf{10.4}            \\
    PAI (\emph{w/} DI) &  \textbf{18.0} & \textbf{5.0} & \textbf{40.2} & 11.2     \\
    \midrule
    VCD  & 25.0 & 8.3 & 59.8 & 17.8          \\
    VCD (\emph{w/} DI)  & \textbf{22.6}  & \textbf{7.4}  & \textbf{45.0} & \textbf{12.1}      \\
    \bottomrule
  \end{tabular}
  \caption{\label{tab:NegContext}
     CHAIR$_{64}$ and CHAIR$_{512}$ represent results on the CHAIR benchmark with maximum generation lengths of 64 and 512, respectively. (\emph{w/} DI) indicates the method with our proposed distinct contrastive inputs.}
\end{table}

\paragraph{Distorted Contrastive Distributions.} We analyze the effect of distorted distributions resulting from contrastive visual inputs. As previously discussed, methods such as VCD~\cite{VCD} and PAI~\cite{PAI} introduce a training–inference gap due to the use of noisy or absent visual inputs. To examine whether these distorted distributions negatively impact the performance of contrastive decoding, we directly apply our contrastive input construction method to both VCD and PAI. The results, shown in Table~\ref{tab:NegContext}, indicate that the performance of both methods improves when using our construction approach. This confirms that distorted contrastive distributions can indeed degrade the effectiveness of contrastive decoding.

\begin{table*}[ht]
  \centering
  \begin{tabular}{llccccccccc}
    \toprule
    \multirow{2}{*}{\textbf{Model}} & \multirow{2}{*}{\textbf{Method}} & \multicolumn{3}{c}{\textbf{CHAIR$_{512}$}} &\multicolumn{3}{c}{\textbf{POPE}(F1$\uparrow$)} &\multicolumn{3}{c}{\textbf{DetailCaps}}  \\
    \cmidrule(rl){3-5} \cmidrule(rl){6-8}\cmidrule(rl){9-11}
    &  & \textbf{Cs}$\downarrow$ & \textbf{Ci}$\downarrow$ & \textbf{Recall}$\uparrow$ & \textbf{Ran.}& \textbf{Pop.} & \textbf{Adv.} & \textbf{Cs}$\downarrow$ & \textbf{Ci}$\downarrow$ & \textbf{CAP}$\uparrow$   \\
     \midrule
     \multirow{2}{*}{LLaVA-Next} & Regular &  41.8 & 11.9 & 60.2 &  85.47 &83.14  &79.87   & 38.0 & 10.4  & 61.06  \\
                                & CICD &  \textbf{37.6} & \textbf{10.4} & \textbf{62.7} &   \textbf{88.22}  &\textbf{86.29} &\textbf{82.76}  & \textbf{36.2} & \textbf{8.5}      &  \textbf{62.19}                          \\
    \midrule
    \multirow{2}{*}{Qwen2-VL} & Regular & 53.8 & 10.0 & 69.4 &  88.33 &85.82  &82.43  & 51.3 & 10.6  & 60.04  \\
                                & CICD &  \textbf{50.6} & \textbf{9.4} & \textbf{73.6} &  \textbf{90.67}  &\textbf{88.07} &\textbf{84.53}  & \textbf{48.4}& \textbf{9.9}   & \textbf{61.79}                             \\
    \midrule
    \multirow{2}{*}{Qwen2.5-VL} & Regular & 41.2 & 11.5&  65.7 &  87.83 &85.66 &83.12  & 42.2 & 11.7 & 62.62   \\
                                & CICD & \textbf{39.4} & \textbf{9.6} & \textbf{68.1} &   \textbf{89.95} &\textbf{88.00} &\textbf{84.75}     & \textbf{43.8} & \textbf{10.9}      & \textbf{65.07}                         \\
    \bottomrule
  \end{tabular}
  \caption{\label{tab:more}
      Results on additional LVLMs. CHAIR$_{512}$ denotes results on the CHAIR benchmark with a maximum generation length of 512. \emph{Ran.}, \emph{Pop.}, and \emph{Adv.} denote the Random, Popular, and Adversarial subsets, respectively.
    }
\end{table*}

\begin{figure}[tbp!]
  \centering
  \includegraphics[width=0.9\linewidth]{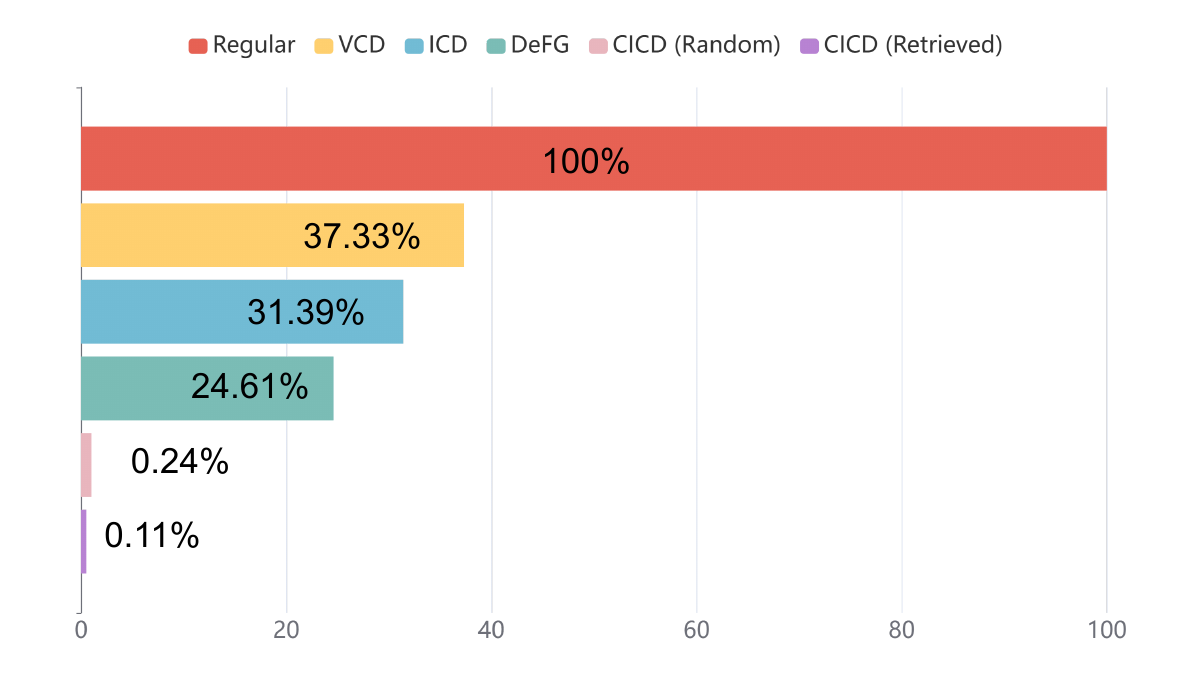}
  \caption{\label{fig:vis_sim}
  Overlap ratio of image-related words between original and negative contexts in contrastive decoding, based on 100 randomly selected COCO images.
}
\end{figure}

\begin{figure}[tbp!]
  \centering
  \includegraphics[width=0.8\linewidth]{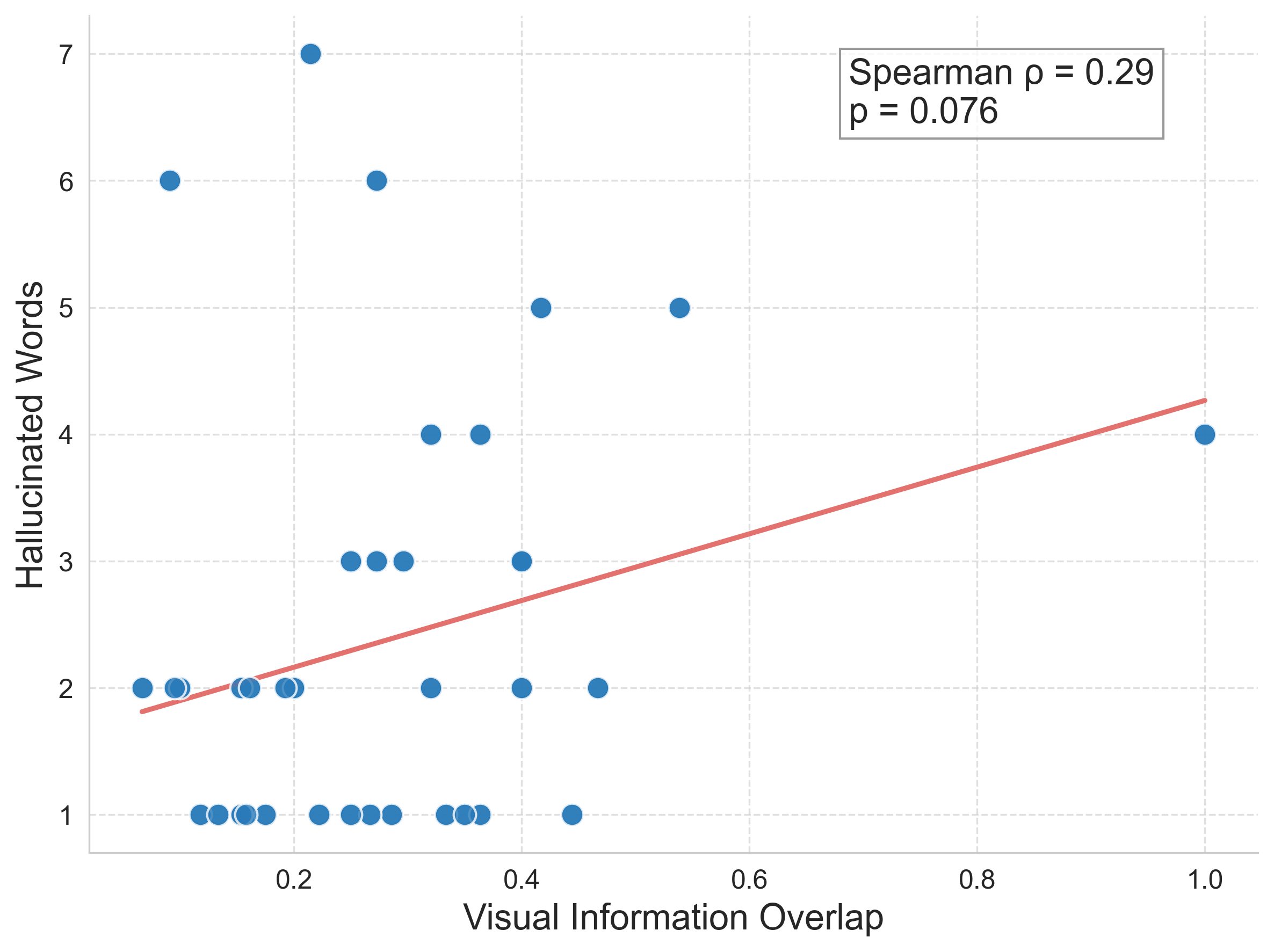}
  \caption{\label{fig:plot}
  Correlation between visual information overlap ratio and hallucination behavior. 
}
\end{figure}

\paragraph{Visual Information Overlap.}
We analyze the visual information overlap between the original and contrastive contexts across different contrastive decoding methods by computing the overlap ratio of image-related words in their descriptions using the MSCOCO dataset. 
Implementation details can be found in Appendix~\ref{sec:Implementation Details for Visual Information Overlap}.
Figure~\ref{fig:vis_sim} presents the overlap ratios for various methods.
The results show that VCD, ICD, and DeGF exhibit substantial overlap with the original input, indicating that their contrastive visual inputs retain a significant amount of shared visual content.
In contrast, our method yields a nearly negligible overlap, suggesting it provides more distinct contrastive contexts.
The information overlap between the two contexts in contrastive decoding confuses the model, amplifying the advantages of suboptimal tokens and thereby inducing hallucinatory behavior.
Furthermore, we analyze the association between overlap ratios and hallucination behaviors based on the DeGF, which has balanced hallucination rates and overlap ratios.
We remove samples without hallucinations and without visual information overlap.
Although the statistical results in Figure~\ref{fig:plot} do not show a strong correlation ($p=0.076$), there is still a positive association ($\rho = 0.29$).
In short, CICD achieves significantly lower hallucination rates compared to VCD, ICD, and DeGF, demonstrating the superiority of low visual overlap in CICD.


\begin{table}[t]
  \centering
    \begin{tabular}{lcccc}
    \toprule
 \multirow{2}{*}{\textbf{Method}} & \multicolumn{2}{c}{\textbf{CHAIR$_{64}$}} & \multicolumn{2}{c}{\textbf{CHAIR$_{512}$}}   \\
    \cmidrule(rl){2-3}  \cmidrule(rl){4-5}
& \textbf{Cs}$\downarrow$ & \textbf{Ci}$\downarrow$ & \textbf{Cs}$\downarrow$ & \textbf{Ci}$\downarrow$  \\
    \midrule
\multicolumn{5}{l}{\emph{Selective Removal of Language Priors}} \\ 
CICD ($\gamma=-4$) & \textbf{18.0} & \textbf{6.1}  &  \textbf{43.8} & \textbf{11.7}      \\
CICD (\emph{w/} DeGF)  & 21.8 & 6.9  & 56.8 & 16.4      \\
\cmidrule(rl){1-5} 
\multicolumn{5}{l}{\emph{Complete Removal of Language Priors}} \\
CICD ($\gamma=-\infty$)   &  19.6 & \textbf{6.1} & 44.4 & \textbf{11.7}    \\
CICD (\emph{w/} VCD)  & 22.6  & 7.4  & 45.0 & 12.1      \\
    \bottomrule
  \end{tabular}
  \caption{\label{tab:DynSupression}
  Comparison among different contrastive decoding methods.
  CICD (\emph{w/} VCD) is equivalent to VCD (\emph{w/} DI).
    }
\end{table}

\paragraph{Excessive Suppression of Language Priors.}
While contrastive decoding is effective in suppressing language priors, it also raises concerns about excessive suppression, which may negatively affect the coherence of generated text.  
To investigate this issue, we set $\gamma$ to $-\infty$ to mitigate all priors and further replace Eq.~\ref{Eq_6} with other contrastive decoding methods to evaluate the superiority of dynamic language prior suppression.
As shown in Table~\ref{tab:DynSupression}, we observe that using cross-image consistency to control the extent of language prior suppression is highly effective, leading to reduced hallucination.  
Moreover, preserving essential priors is crucial, as the performance of CICD degrades when $\gamma$ is set to $-\infty$.
VCD removes all language priors while DeGF augments the retained language priors, both of which yield suboptimal performance.
In particular, DeGF's augmentation of retained priors can harm model performance when prior issues are more severe, such as with longer contexts.
These results indicate that the dynamic language prior suppression method can effectively mitigate language priors without harming the performance of LVLMs.
The impact of the choice of $\gamma$ will be discussed in Appendix~\ref{sec:hyperparameter}.

\subsection{Effectiveness of CICD on More LVLMs}
To comprehensively demonstrate the effectiveness and generalizability of our CICD method, we conduct additional experiments using three of the most recent open-source LVLMs. The results, presented in Table~\ref{tab:more}, show that our method consistently achieves significant performance improvements across all experiments. CICD effectively reduces hallucinations in both generative and discriminative tasks by mitigating the influence of language priors. These results further confirm the transferability of CICD to a variety of LVLMs.

\section{Conclusion}
In this paper, we first reveal that language priors are intrinsic to LVLMs and remain consistent across different images. Building on this observation, we propose Cross-Image Contrastive Decoding (CICD), a training-free method that effectively suppresses language priors without compromising the model's performance. Despite its simplicity and efficiency, CICD demonstrates strong practical potential and can be seamlessly integrated with other hallucination mitigation methods to further enhance LVLM performance.

\section*{Acknowledgments}
This work is supported by Zhongguancun Academy Project No.20240103.

\bibliography{custom}

\clearpage

\appendix

\section{Limitation}
We observe that language priors remain consistent across different images. Based on this insight, we propose the Cross-Images Contrastive Decoding (CICD) method to eliminate language priors.
Our method has two main limitations compared to regular decoding.
First, it requires an additional image that differs sufficiently from the original input.
Second, both the original and the additional images must undergo separate forward passes to obtain their respective logit distributions for contrastive decoding.
\paragraph{Addtional Image}
Although the additional image is indispensable, its acquisition is straightforward.
Hallucinations typically occur in content-rich elements such as objects, attributes, and relationships, which naturally exhibit substantial variability across real-world images.
We demonstrate that random selection and retrieval-based methods achieve comparable low visual information overlap ratios when selecting images.
The experimental results also show that randomly selected images perform comparably to retrieved ones, highlighting the practical applicability of our method.
This flexibility significantly lowers the barrier for real-world deployment.

\paragraph{Twice Forward Propagation}
Although performing twice forward propagations doubles the theoretical computational cost, in practice, we merge the two inputs into a single batch and perform one forward propagation with a doubled batch size to achieve the same effect.
Under this implementation, our method incurs only a 10\% increase in inference time and a 5\% increase in memory usage, demonstrating strong potential for practical deployment.

\begin{table*}[htb]
  \centering
  \begin{tabular}{llcccccccc}
    \toprule
    \multirow{2}{*}{\textbf{MaxLen}} & \multirow{2}{*}{\textbf{Method}} & \multicolumn{4}{c}{LLaVA-1.5} & \multicolumn{4}{c}{InstructBLIP}  \\
    \cmidrule(rl){3-6} \cmidrule(rl){7-10}
    & & \textbf{Cs}$\downarrow$ & \textbf{Ci}$\downarrow$ & \textbf{Recall} & \textbf{Length} & \textbf{Cs }$\downarrow$ & \textbf{Ci}$\downarrow$ & \textbf{Recall} & \textbf{Length}  \\
    \midrule
    \multirow{7}{*}{64} & Regular & 24.4&8.9&56.6&53.7 & 35.6&13.2&56.4&54.8    \\
                            & VCD  & 25.0&8.3&\underline{59.0}&54.0 &  32.2&10.3&\underline{60.6}&55.2                            \\
                            & ICD &  23.2&8.1&58.4&54.4 & 29.8&9.8&\underline{60.6}&55.7                             \\
                            & IBD &  21.2&6.9&58.8&54.5 & 27.8&9.2&60.3&56.1   \\
                            & PAI &  \underline{20.0} & \underline{6.2} & 56.9 & 55.1 & \underline{26.0} & \underline{8.9} & 53.9 & 54.4 \\
                            & DeGF &  22.4&7.2&58.2&54.4 & 32.4&11&59.4&56.0                            \\
                      \cmidrule(rl){2-10}
                      & CICD & \textbf{18.0}&\textbf{6.1}&\textbf{59.6}&53.3 & \textbf{23.8}&\textbf{7.7}&\textbf{62.2}&54.8       \\
     \midrule
      \multirow{7}{*}{512} & Regular & \underline{54.6}&16.4&72.6&106.8 & 62.6&19.5&66.9&113.6     \\
      & VCD  & 59.8&17.8&\textbf{75.6}&104  & 64.8&18.8&\textbf{71.9}&108.6                      \\
      &  ICD &  57.0&\underline{15.0}&74.6&103.3  & 59.0&17.1&69.2&108.5               \\
      & IBD &  57.6&16.5&74.2&104.3&\underline{57.6}&\underline{15.7}&70.8&118.5          \\
      & PAI &    \textbf{41.2} & \textbf{10.4} & 68.6 & 116.2 & 67.6 & 19.4 & 68.0 & 143.1          \\
     &   DeGF &  57.4&16.3&\underline{75.5}&102.8  & 59.0&17.7&\underline{71.5}&108.9     \\
       \cmidrule(rl){2-10}
     &    CICD & \underline{43.8}&\underline{11.7}&75.0&103.3 & \textbf{49.8}&\textbf{13.7}&70.3&98.8    \\
    \bottomrule
  \end{tabular}
  \caption{
      Results on the test set of CHAIR.
      \emph{MaxLen} denotes the maximum generation length.
      \emph{Cs} and \emph{Ci} represent
      \emph{CHAIRs} and \emph{CHAIRi} respectively.
      }\label{tab:chair}
\end{table*}

\begin{table*}[htb]
  \centering
  \setlength{\tabcolsep}{1mm}
  \begin{tabular}{llccccccccc}
    \toprule
    \multirow{2}{*}{\textbf{Model}} & \multirow{2}{*}{\textbf{Method}} & \multicolumn{4}{c}{\textbf{Generative}} & \multicolumn{4}{c}{\textbf{Discriminative}} &  \\
    \cmidrule(rl){3-6} \cmidrule(rl){7-10}
    &  & \textbf{CHAIR}$\downarrow$ & \textbf{Cover}$\uparrow$ & \textbf{Hal}$\downarrow$ & \textbf{Cog} $\downarrow$ & \textbf{Acc.}$\uparrow$ & \textbf{Prec.}$\uparrow$ & \textbf{Recall}$\uparrow$ & \textbf{F1} $\uparrow$ & \textbf{AMBER} $\uparrow$   \\
     \midrule
     \multirow{7}{*}{LLaVA-1.5} & Regular &  11.6&49.7&47.7&4.4&67.4&83.9&61.8&71.2&79.80     \\
                      & VCD  &  9.8&\underline{51.2}&43.8&4.4&68.1&85.1&61.0&71.1&80.65                      \\
                     & ICD &   8.8&\underline{51.2}&38.7&4.1&70.3&86.0&\underline{64.1}&\underline{73.4}&82.30                    \\
                     & IBD &   9.8&50.5&42.2&4.4&69.2&86.3&62.1&72.2&81.20                     \\
                     & PAI &   \underline{7.7} & 49.3 & \underline{36.9} & \underline{3.3} & \underline{70.5} & 86.5 & \textbf{64.2} & \textbf{73.7} & \underline{83.0}         \\
                      & DeGF &  9.1&50.7&39.9&4.1&70.2&\underline{86.8}&63.0&73.0&81.95                      \\
                      \cmidrule(rl){2-11}
                      & CICD &  \textbf{6.6}&\textbf{52.7}&\textbf{34.8}&\textbf{2.2}&\textbf{71.1}&\textbf{89.4}&61.9&73.1&\textbf{83.25}                     \\
    \midrule
    \multirow{7}{*}{InstructBLIP} & Regular & 12.4&51.9&52.4&5.0&68.3&78.6&71.7&75&81.30    \\
                            & VCD  &    9.9&54.0&44.6&\underline{4.2}&70.6&80.6&\underline{72.7}&76.4&83.25                            \\
                            & ICD & 9.8&53.9&46.7&5.1&69.9&80.3&70.9&75.3&82.75                            \\
                            & IBD &  \underline{9.0}&\textbf{56.1}&45.1&4.6&54.2&\textbf{86.3}&34.2&49.0&70.00                            \\
                            & PAI &   11.7 & 52.8 & 55.1 & 5.4 & 68.0 & 78.3 & 67.0 & 72.2 & 80.25                     \\
                            & DeGF &  9.7&\underline{54.1}&\underline{44.5}&5.2&\underline{72.1}&80.7&\textbf{75.1}&\textbf{77.8}&\underline{84.05}                            \\
                      \cmidrule(rl){2-11}
                      & CICD & \textbf{7.1}&53.6&\textbf{35.0}&\textbf{2.3}&\textbf{72.8}&\underline{85.4}&70.1&\underline{77.0}&\textbf{84.95}                            \\
    \bottomrule
  \end{tabular}
  \caption{
      Results on AMBER. The maximum generation length is 512.
      \emph{AMBER} metirc is calculated by $(100-CHAIR+F1)/2$.
    }\label{tab:amber}
\end{table*}

\section{Implementation Details for Baselines}\label{sec:Implementation Details for Baselines}
The implementation details for the reproduction of baselines are as follows:
\begin{itemize}
    \item \textbf{VCD}~\cite{VCD}: We added 500 steps of Gaussian noise to the images to construct the negative visual input for contrastive decoding. The hyperparameters in VCD are set to $\alpha=1, \beta=0.1$.
    \item \textbf{ICD}~\cite{ICD}: We use the sentence \emph{"You are a confused objects detector to provide a fuzzy overview or impression of the image."} as the disturbance instruction for contrastive decoding. The hyperparameters for contrastive decoding are set according to VCD.
    \item \textbf{IBD}~\cite{IBD}: We follow the practice in the original paper, setting $\epsilon=2$, $\alpha=1.5\times 10^4$, and $\beta=0.1$ in our experiments.
    \item \textbf{PAI}~\cite{PAI}: We set the visual attention augmentation coefficient $\alpha$ to 0.5 and enable increased visual attention starting from layer 2 (counting from 0). The hyperparameter $\gamma$ is set to 1.1. PAI employs adaptive plausibility constraints with $\beta=0.1$.
    \item \textbf{DeGF}~\cite{DeGF} We employ LLaVA-v1.5-7B to generate captions for each image and use Stable Diffusion v1-5 to generate images from the captions. The parameter $\gamma$ is set to 0.1, and $\beta$ is set to 0.1 for the image captioning task and 0.25 for the visual question answering task.
\end{itemize}
\textbf{Note: The same parameters may have different meanings across methods. For specific details, please refer to the original papers.}

\section{Implementation Details for Visual Information Overlap}\label{sec:Implementation Details for Visual Information Overlap}
First, we apply several contrastive decoding methods, each constructing its own negative context, and use them to prompt LVLMs to answer user queries. For comparison, we obtain standard responses via regular decoding. We then use QwenVL-Max-0125 to extract image-related words from the generated responses. Inspired by CoT~\cite{cot}, we guide the model through a multi-stage extraction process: it first identifies all visually relevant words without access to the image, and then selects those strongly associated with the image content, which we define as \textit{relevant words}.

\begin{figure*}[htbp]
  \centering
  \includegraphics[width=\textwidth]{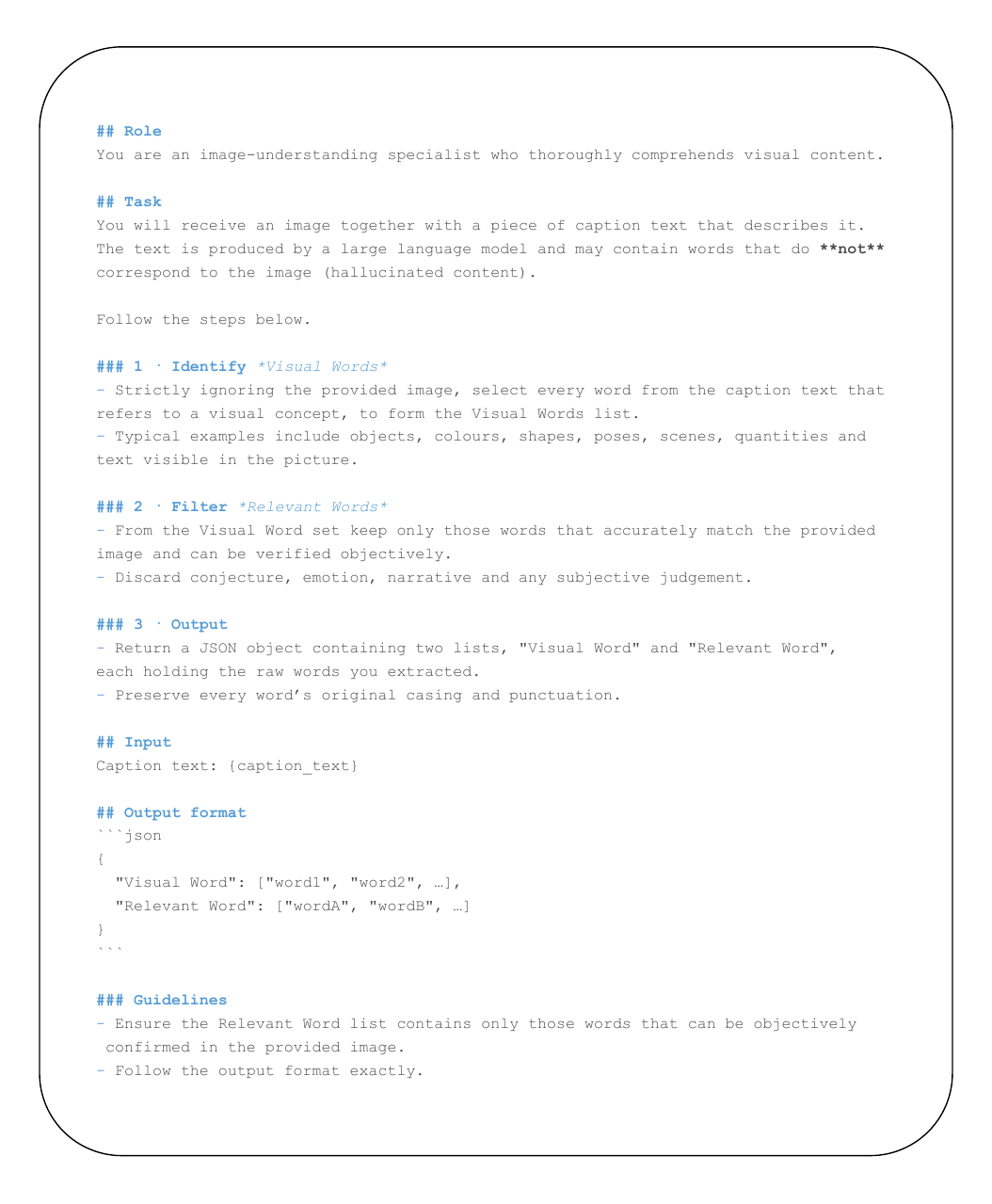}
  \caption{Prompt for guiding QwenVL-Max to progressively extract relevant words from the response.}
  \label{fig:prompt}
\end{figure*}

Finally, we compute the overlap ratio between the relevant words generated from negative contexts and those obtained through regular decoding. This ratio quantifies how much essential visual content is preserved in responses generated from negative contexts, serving as an indicator of retained visual information. The prompt used to guide QwenVL-Max is shown in Figure~\ref{fig:prompt}.

\begin{table*}[htbp]
  \centering
  \renewcommand{\arraystretch}{1}
  \setlength{\tabcolsep}{6pt}
  \begin{tabular}{llcccccc}
    \toprule
    \multirow{2}{*}{\textbf{Model}} & \multirow{2}{*}{\textbf{Method}} & \multicolumn{3}{c}{\textbf{CHAIR}(MaxLen=512)} &\multicolumn{3}{c}{\textbf{POPE }(F1$\uparrow$)}  \\
    \cmidrule(rl){3-5} \cmidrule(rl){6-8}
    &  & \textbf{Cs}$\downarrow$ & \textbf{Ci}$\downarrow$ & \textbf{Recall} & \textbf{Random}& \textbf{Popular} & \textbf{Adversarial}   \\
    \midrule
     \multirow{2}{*}{LLaVA-1.5-13B} & Regular & 58.8 & 17.0 & 73.4 &  82.94 & 81.46 & 78.49      \\
                                & CICD &  \textbf{44.4} & \textbf{12.9} & \textbf{76.0}  &  \textbf{87.92} &\textbf{85.68}&\textbf{80.99}                                   \\
    \midrule
     \multirow{2}{*}{INstructBLIP-13B} & Regular & 65.6 & 21.0 & 62.7 &  81.31 & 77.20 & 74.69       \\
                                & CICD &  \textbf{44.0} & \textbf{11.6} & \textbf{65.7}  & \textbf{87.52} & \textbf{83.60} & \textbf{78.96}    \\
    \bottomrule
  \end{tabular}

  \caption{
      Results on larger LVLMs. The maximum generation length is 512.
    }\label{tab:larger}
\end{table*}

\section{More Experimental Results}\label{sec:Detailed Experiment}
\subsection{Benchmarks}
In order to validate the effectiveness of our method in mitigating language hallucination, we evaluate its performance on four widely-used multimodal hallucination benchmarks: two generative benchmarks (CHAIR \cite{chair} and DetailCaps \cite{DetailCap}), one discriminative benchmark (POPE \cite{POPE}), and one hybrid benchmark (AMBER \cite{AMBER}).

\paragraph{CHAIR} evaluates the proportion of hallucinated objects, which are generated by the model but not present in the reference annotations. Following prior works, we randomly select 500 images from the MSCOCO \cite{coco} dataset as the test set. We additionally select another 500 images to construct a validation set for hyperparameter tuning. 
This benchmark includes two metrics: CHAIRs and CHAIRi, defined as follows:
\begin{equation}
\begin{aligned}
   \text{CHAIRs} &= \frac{|\text{Hallucinated Objects}|}{|\text{All Objects}|}, \\
   \text{CHAIRi} &= \frac{|\text{Hallucinated Captions}|}{|\text{All Captions}|}
\end{aligned}
\end{equation}

\paragraph{DetailCaps} is a fine-grained image captioning benchmark, accompanied by ground-truth detail captions generated by GPT-4V \cite{GPT-4v}, Gemini-1.5-Pro \cite{Gemini-15}, and GPT-4o \cite{GPT-4o} for evaluation. It contains 4,870 images collected from multiple datasets, and we use a subset of 700 images from MSCOCO to conduct our experiments. 
This benchmark primarily adopts the CAPTURE metric to evaluate caption quality.
CAPTURE evaluates the alignment between generated and reference captions by calculating F1 scores using both hard and soft matching across entities ($F1_{obj}$), attributes ($F1_{attr}$), and relations ($F1_{rel}$). The final score is computed through a weighted aggregation of these components:
\begin{equation}
   CAPTURE = \frac{\alpha F1_{obj}+\beta F1_{attr}+\gamma F1_{rel}}{\alpha+\beta+\gamma}
\end{equation}
where $\alpha=5,\beta=5,\gamma=2$

\paragraph{POPE} is a widely adopted benchmark for evaluating object hallucinations by prompting LVLMs to identify whether a specific object is present in the image. It comprises three distinct datasets: \emph{MSCOCO}, \emph{A-OKVQA}, and \emph{GQA}.
Each dataset uses three different negative sampling settings: \emph{Random}, \emph{Popular}, and \emph{Adversarial}.
In total, it includes 9,000 questions and 1,500 images. Accuracy and F1 score are used as the primary evaluation metrics.

\paragraph{AMBER} combines generative and discriminative tasks, and is evaluated on a curated set of 1,004 images. In addition to image captioning, it includes 14,216 questions designed to assess hallucinations in object, attribute, and relation recognition.
AMBER contians mulitple metrics: \emph{CHAIR}, \emph{Cover}, \emph{Hal}, \emph{Cog}. 
It provides an annotated objects list $A_{obj}={obj_1^A, obj_2^A, \cdots, obj_n^A}$, and the generated objects are labeled as $R'_{obj}$.
Each metric is calculated as follows:
\begin{equation}
\begin{aligned}
   \text{CHAIR} &= 1 - \frac{\mathrm{len}(R'_{obj} \cap A_{obj})}{\mathrm{len}(R'_{obj})},\\
   \text{Cover} &= \frac{\mathrm{len}(R'_{obj} \cap A_{obj})}{\mathrm{len}(A_{obj})}, \\
   \text{Hal}   &= \frac{\{ \text{CHAIR} > 0 \}}{\{\text{All Caps}\}}, \\
   \text{Cog}   &= \frac{\mathrm{len}(R'_{obj} \cap H_{obj})}{\mathrm{len}(R'_{obj})},
\end{aligned}
\end{equation}
where $H_{obj}$ denotes the set of hallucinated target objects generated by the LVLMs, and \emph{All Caps} refers to all generated captions.

\subsection{Evaluated LVLMs}
To demonstrate the generalizability of the proposed CICD as a broadly applicable plug-and-play module, we evaluate its performance across six different LVLMs spanning three model families: InstructBLIP~\cite{blip}, LLaVA-1.5 \cite{llava15}, LLaVA-Next \cite{llava-next}, Qwen-VL-Chat \cite{qwen-vl}, Qwen2-VL-Instruct \cite{qwen2-vl}, and Qwen2.5-VL-Instruct \cite{qwen25-vl}. InstructBLIP and LLaVA-1.5 are both built upon Vicuna-7B \cite{vicuna} as their language backbone, while LLaVA-Next is based on LLaMA3. The Qwen-VL series, including Qwen2-VL-Instruct and Qwen2.5-VL-Instruct, is developed using the QwenLM framework. As for vision encoders, InstructBLIP, LLaVA-1.5, LLaVA-Next, and Qwen-VL-Chat utilize the pretrained CLIP model~\cite{CLIP}, whereas Qwen2-VL-Instruct and Qwen2.5-VL-Instruct adopt a fine-tuned ViT \cite{ViT}. All models are implemented at the 8B parameter scale.

\subsection{Results on CHAIR}\label{sec:Results on CHAIR}

The results in Table~\ref{tab:chair} demonstrate our method's superior performance across all experimental conditions, with particularly pronounced improvements observed at a longer generation length.
Longer context lengths amplify the model's susceptibility to language priors, making these experimental results particularly compelling evidence of our method's efficacy in mitigating such priors.
Compared to baselines, our method's principal advantage lies in its complete preservation of visual information and essential language priors, which is directly reflected in the superior performance in image captioning.

\subsection{Results on AMBER}\label{sec:Results on AMBER}
As illustrated in Table~\ref{tab:amber}, our method achieves the lowest hallucination frequency in generative tasks while simultaneously attaining state-of-the-art performance on the comprehensive AMBER metric.
Language prior is particularly critical for generative tasks, where language priors manifest more severely than in discriminative tasks, explaining our method's pronounced advantages in image captioning.

\subsection{Effectiveness on Larger LVLMs}
We evaluate the effectiveness of our method on larger models using two benchmarks, POPE and CHAIR, with the results presented in Table\ref{tab:larger}.
Our method continues to deliver significant performance improvements on a 13B model, indicating its strong generalizability.
These results suggest that our approach scales well and remains effective across different model sizes.

\subsection{Effects of $\gamma$}\label{sec:hyperparameter}

While language priors contribute to hallucinations in LVLMs,
they also play a critical role in maintaining textual coherence and fluency.
We employ JS divergence to selectively suppress language priors, where priors exhibiting high cross-image consistency are retained, as reflected by lower JS divergence scores.
We establish a threshold of $\gamma = -4$ in Eq. 2, based on the statistical distribution of JS divergence scores.

\begin{figure}[htbp]
  \centering
  \includegraphics[width=0.8\linewidth]{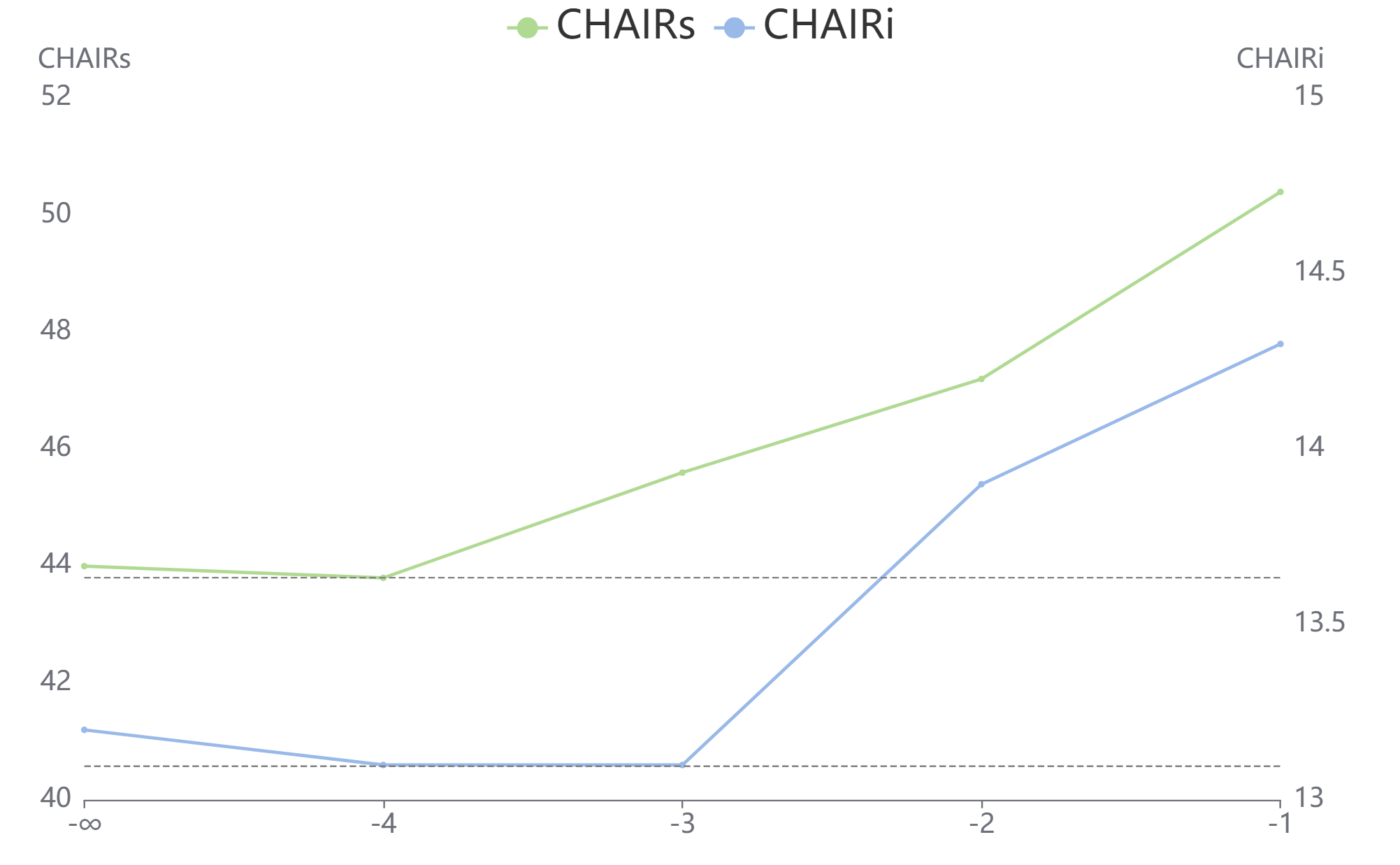}
  \caption{
  The results on the validation set of CHAIR using LLaVA-1.5.
  The X-axis represents the value of $\gamma$.
  $\gamma=-\infty$ is equivalent to removing both detrimental and essential priors.
  }
  \label{Fig_jsd_th}
\end{figure}

A comprehensive analysis is shown in Figure~\ref{Fig_jsd_th}.
The experimental results align with the statistical analysis, demonstrating optimal performance at $\gamma = -4$.
Another notable finding is that preserving necessary priors is beneficial, as it prevents the model's generative capability from being compromised during the contrastive decoding process, thereby further reducing hallucinations.

\begin{figure*}[htbp]
  \centering
  \includegraphics[width=\textwidth]{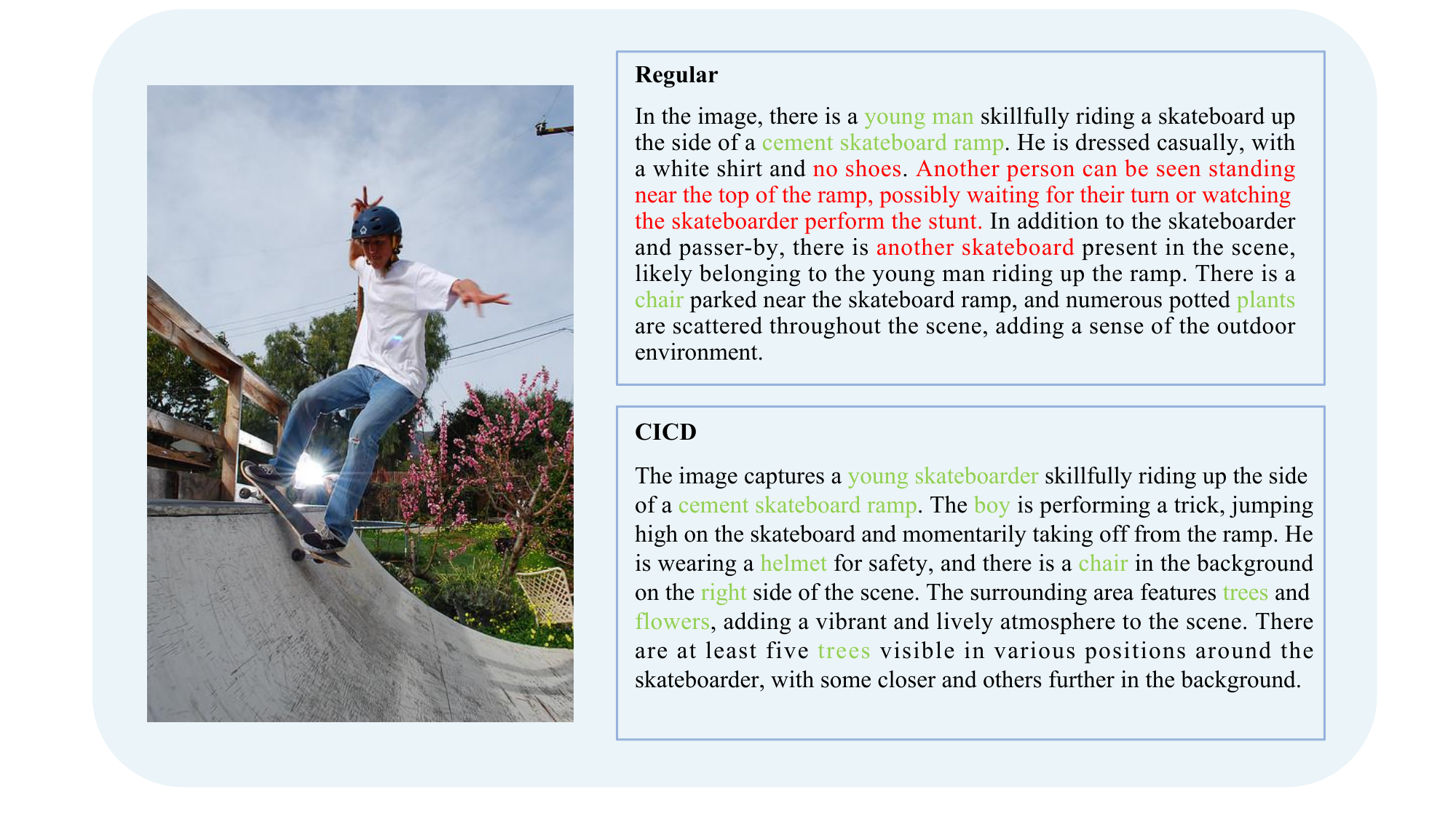}
  \caption{Case study (1).}
  \label{fig:case1}
\end{figure*}
\begin{figure*}[htbp]
  \centering
  \includegraphics[width=\textwidth]{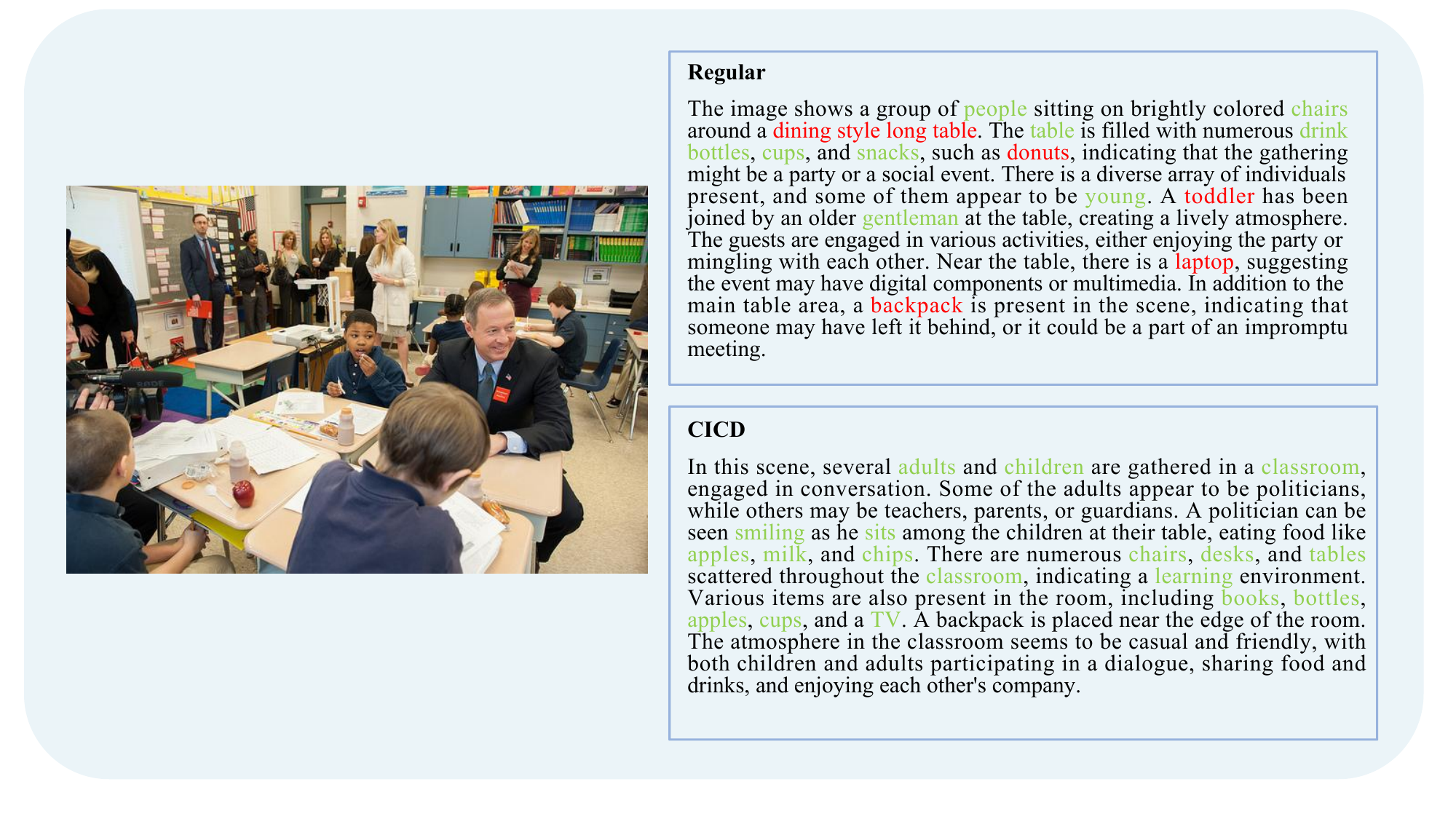}
  \caption{
  Case study (2).
  }
  \label{fig:case2}
\end{figure*}

\section{Case Study}
We present case studies to illustrate the superior performance of our method in hallucination reduction, as shown in Figure~\ref{fig:case1} and Figure~\ref{fig:case2}. The image captions are generated by LLaVA-1.5. Red highlights indicate hallucinated content, while green highlights denote content consistent with the image.

We can clearly observe that Regular Decoding produces a substantial amount of hallucinated content, often manifesting as content words such as objects, attributes, and relationships. In Figure~\ref{fig:case1}, for example, although the image depicts a skateboarding scene, the caption includes a reference to "spectators"—a detail not present in the image. Similarly, Figure~\ref{fig:case2} shows an eating scene, yet the caption introduces a "long dining table," which is not actually visible. These hallucinations stem from LVLMs inheriting prior knowledge from LLMs, leading to biased descriptions in certain typical scenarios. In contrast, our method significantly reduces such hallucinations. 

Overall, our method effectively mitigates the influence of language priors by leveraging cross-image consistency.

\end{document}